\documentclass[letterpaper, 10 pt, conference]{ieeeconf}  

\IEEEoverridecommandlockouts                              
\usepackage{graphicx}
\usepackage{url}
\usepackage{pifont}

\usepackage{amsmath}

\title{\LARGE \bf
Landslide Detection and Segmentation Using\\Remote Sensing Images and Deep Neural Network
}

\author{Cam~Le$^{1*}$, 
        Lam~Pham$^{1*}$,
        Jasmin~Lampert$^{1}$,
        Matthias~Schlögl$^{2,3}$, 
        Alexander~Schindler$^{1}$
\thanks{C. Le, L. Pham, J. Lampert, and A. Schindler are with the Competence Unit Data Science \& Artificial Intelligence at the Austrian Institute of Technology, Austria.}%
\thanks{M. Schlögl is with the Department for Climate Impact Research at
GeoSphere Austria and the Institute of Mountain Risk Engineering at the
University of Natural Resources and Life Sciences, Vienna, Austria}%
\thanks{(*) Main and equal contribution into the paper.}
}

\begin{document}

\maketitle
\thispagestyle{empty}
\pagestyle{empty}

\begin{abstract}
Knowledge about historic landslide event occurrence is important for supporting disaster risk reduction strategies. Building upon findings from 2022 Landslide4Sense Competition, we propose a deep neural network based system for landslide detection and segmentation from multisource remote sensing image input.
We use a U-Net trained with Cross Entropy loss as baseline model.
We then improve the U-Net baseline model by leveraging a wide range of deep learning techniques.
In particular, we conduct feature engineering by generating new band data from the original bands, which helps to enhance the quality of remote sensing image input.
Regarding the network architecture, we replace traditional convolutional layers in the U-Net baseline by a residual-convolutional layer.
We also propose an attention layer which leverages the multi-head attention scheme.
Additionally, we generate multiple output masks with three different resolutions, which creates an ensemble of three outputs in the inference process to enhance the performance.
Finally, we propose a combined loss function which leverages Focal loss and IoU loss to train the network.
Our experiments on the development set of the Landslide4Sense challenge achieve an F1 score and an mIoU score of 84.07 and 76.07, respectively.
Our best model setup outperforms the challenge baseline and the proposed U-Net baseline, improving the F1 score/mIoU score by 6.8/7.4 and 10.5/8.8, respectively.

\indent \textit{Items}--- Convolutional neural network, landslide, remote sensing image. 
\end{abstract}


\section{INTRODUCTION}
\label{intro}
Natural hazards pose a severe threat to the lives of people around the world. In particular, landslides are a major cause of losses in mountainous areas~\cite{picarelli2021impact, emdat}. 
Knowledge about historic landslide event occurrence is of core importance in the context of quantitative risk assessment, which in turn supports the design and implementation of effective disaster risk reduction strategies.
Several methodological approaches are used for detecting and mapping different types of landslides. In addition to manual visual interpretation, different automated methods that leverage different types of data sets have been developed. Most notably, these methods include
the analysis of digital terrain models derived through airborne laserscanning, e.g. by using geographic object-based image analysis~\cite{Knevels2019} or LiDAR altimetry~\cite{mckean2004objective}, the analysis of aerial photographs \cite{Li2016}, or various change detection methods applied to multi-spectral or SAR data~\cite{Handwerger2021, Plank2016}.


While these methods are tried and tested, the rapid technological development in intersection of remote sensing imagery and image segmentation using increasingly advanced neural network architectures has opened up new possibilities for landslide detection and mapping compared to the conventional methods~\cite{garcia2023relict}. 
The availability of free multi-spectral remote sensing imagery from the satellites, combined with advances in computer vision and machine learning, enables the development of automated landslide detection and segmentation frameworks at comparably low cost.

Recent attempts at developing such systems, which are based on deep neural network architectures such as U-Net, DeepLab, Transformers \cite{huang2023landslide, fu2021improved} or on adapted pre-trained models such as variants of ResNet or EfficientNet \cite{liu2023feature, tang2021mill}, have presented very promising results.  
Most of the published systems were based on dedicated datasets collected by the authors~\cite{soares2022landslide, zhang2019characteristics} or onsynthetic datasets~\cite{zhou2022novel}.
As a result, these datasets only reflect the landslide events of a certain region, which leads to certain limitations in the developed landslide detection systems.
%

The Landslide4Sense dataset published by Ghorbanzadeh et al. in 2022~\cite{ls_data, Ghorbanzadeh2022} constitutes an interesting and large dataset aimed at landslide detection and segmentation. The data set mainly consists of  multi-spectral remote sensing images from Sentinel-2 and (presumably) elevation information as used in the ALOS PALSAR RTC products (i.e., SRTM and NED DEM with geoid correction applied)\footnote{This information is not really clear from Ghorbanzadeh et al. (2022), who misleadingly state that "DEM and slope data from ALOS PALSAR"~\cite{ls_data} were used.}.


Based on the Landslide4Sense dataset, we present a deep neural network based system for landslide detection and segmentation, including the following specific improvements over the benchmark results \cite{ls_data}:
\begin{itemize}
    \item We first conducted an analysis on how to improve the quality of input remote sensing images by using multiple techniques of data augmentation (random rotation, cutmix) and feature engineering techniques (RGB normalization, feature combination, Gaussian filters, gradient image, Canny Edge detector). 
    \item Second, we improve of the U-Net architecture by proposing a residual-convolutional layer and an attention layer.
    \item Third, we propose a combination of multi-resolution segmentation heads with multiple loss functions, which also helps to improve model performance. 
\end{itemize}

\section{Dataset and methodological background}
\label{dataset}

\subsection{Landslide4Sense dataset}

The benchmark Landslide4Sense dataset~\cite{ls_data} comprises three main subsets: the development set, the evaluation set, and the test set.
While the development set was published with the labels, no labels have been provided for both evaluation set and test set as these subsets were used for the competition challenge~\cite{data_web}.
Therefore, only the development set of the Landslide4Sense dataset is considered in this paper.
This development set comprises 3799 multi-spectral images which were collected from the open source Sentinel-2~\cite{sentinel2} and supplemented with information from ALOS PALSAR.
Each of multi-spectral image presents 14 bands: multi-spectral data from Sentinel-2 (B1, B2, B3, B4, B5, B6, B7, B8, B9, B10, B11, B12); slope data from ALOS PALSAR (B13); and elevation data (DEM) from ALOS PALSAR (B14).
All bands in the dataset have an image size of 128$\times$128. The original spatial resolution of the single bands varies according to the resolution of the source spectral bands of the MSI aboard Sentinel-2: B1, B9 and B10's resolution is 60m, B2 to B4 and B8 were captured at a resolution of 10m per pixel, and B5-B7, B11 and B12 have a resolution of 20m. 
As a result, each of multi-spectral images is an array of shape 128$\times$128$\times$14.
One multi-spectral image of 128$\times$128$\times$14 comes with a label image of 128$\times$128, referred to as the ground truth mask.
The ground truth mask presents a binary image in which landslide pixels and non-landslide pixels are marked by one and zero values, respectively.
Although approximately 58\% of images in the Landslide4Sense development set contains landslide labels, the landslide pixels are minority with only 2.3\% of all pixels being labeled as events. 
Additionally, the ratio of landslide pixels over an image presents a wide range of values from 0.0061\% (i.e., only one pixel out of 128$\times$128 pixels in one image) to 47.53\% (i.e., nearly a half of pixels in an image).
As a result, the dataset presents an imbalance between landslide and non-landslide pixels which causes challenges in the segmentation task.

\subsection{Task definition}

Using the development set of the Landslide4Sense~\cite{ls_data} dataset as a basis, two tasks of landslide detection and landslide segmentation using deep neural network are proposed in this paper\footnote{The segmentation task was not part of the Landslide4Sense challenge~\cite{chal_baseline}.}.
We evaluate our proposed deep neural networks 
using random train-test splitting, using 80\% for training and 20\% as holdout for testing.
When the best configuration of the deep neural network is indicated, we evaluate the best network with the 5-fold cross-validation.
The final evaluation scores are obtained using the average of scores from 5 folds.

\subsection{Evaluation metrics}
Following the guidelines of the Landslide4Sense challenge, we use the F1 score as main evaluation metric~\cite{ls_data, data_web}. In addition, we report the mean Intersection over Union (mIoU) score, which is a crucial performance metric in the segmentation task~\cite{iou_metric}.

%
\section{Proposed U-Net baseline}
\label{baseline}
The baseline model for for landslide detection and segmentation comprises two main components: Online data augmentation and U-Net based network architecture.

\subsection{Online data augmentation}
\label{augmentation}
For the baseline model, we apply two data augmentation methods, rotation and cutmix, to the image input of size 128$\times$128$\times$14. 
We first randomly rotate each image using an angle of 90, 180, or 270 degrees to generate a new image, referred to as the rotation.
Subsequently, random landslide regions from 0 to 2 random landslide images are cut and mixed with the current processing image, referred to as the cutmix \cite{cutmix_image}.
As these data augmentation methods are conducted on each batch of images during training, refer to as step as "online data augmentation".

\subsection{Proposed U-Net based baseline architecture}
\label{baseline_net}
\begin{figure}[ht]
    	\vspace{-0.2cm}
    \centering
    \includegraphics[width =1.0\linewidth]{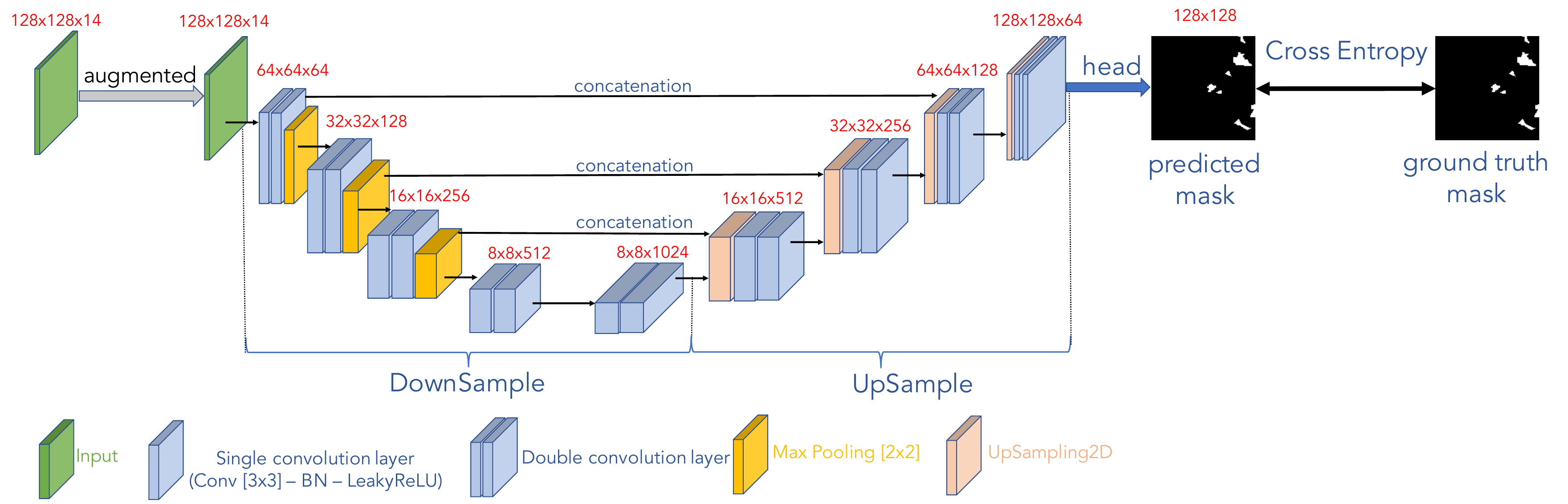}
	\caption{The U-Net baseline architecture.}
    \label{fig:f11}
   \vspace{-0.3cm}
\end{figure}

The proposed baseline leverages a U-Net architecture (Table~\ref{table:baseline}, Fig.~\ref{fig:f11}).

%
The U-Net baseline comprises three main blocks: downsample, upsample, and head.
Both downsample and upsample blocks make use of the same double convolution layer.
The double convolution layer comprises two single convolution layers, each of which contains one convolutional layer (Conv[3$\times$3]), one Batch Normalization layer (BN)~\cite{batchnorm}, and one Leaky Rectified Linear Unit layer (LeakyReLU)~\cite{leak_relu}), as shown in Fig.~\ref{fig:f11}.
While the downsample block scales down the input images of 128$\times$128$\times$14 to 8$\times$8$\times$1024 by using the Max Pooling layer (MP[2$\times$2]), the upsample block scales up the output of downsample block to 128$\times$128$\times$64 by applying UpSampling 2D.
%
The head block, which uses one dropout layer, one convolutional layer (Conv[1$\times$1]) and applies a Softmax function, helps to transform the output of upsample block to the image of $128{\times}128$, referred to as the predicted mask.
The predicted mask is compared with the ground truth mask using Cross Entropy as loss function.

\begin{table}[ht]
    \caption{The U-Net baseline architecture} 
    \vspace{-0.2cm}
    \centering
    \scalebox{0.85}{
    \begin{tabular}{lll} 
        \hline 
        \textbf{Blocks} & \textbf{Sub-blocks \& Layers}   &  \textbf{Output}  \\
        \hline 
        & Input & $128{\times}128{\times}14$ \\
        \hline
         DownSample & Double convolution layer - MP layer[$2{\times}2$]  & $64{\times}64{\times}64$\\
                    & Double convolution layer - MP layer[$2{\times}2$]  & $32{\times}32{\times}128$\\
                    & Double convolution layer - MP layer[$2{\times}2$]  & $16{\times}16{\times}256$\\
                    & Double convolution layer - MP layer[$2{\times}2$]  & $8{\times}8{\times}512$\\
                    & Double convolution layer - Single convolution layer  & $8{\times}8{\times}1024$\\
        \hline
         UpSample & Upsampling2D layer - Double convolution layer & $16{\times}16{\times}512$\\
                  & Upsampling2D layer - Double convolution layer & $32{\times}32{\times}256$\\
                  & Upsampling2D layer - Double convolution layer & $64{\times}64{\times}128$\\
                  & Upsampling2D layer - Double convolution layer & $128{\times}128{\times}64$\\
         \hline
         Head     & Dropout layer(0.2) - Conv layer[1x1] - Softmax & $128{\times}128$ \\
         \hline
    \end{tabular}
    }
    \vspace{-0.2cm}
    \label{table:baseline} 
\end{table}
We construct the U-Net baseline with the Tensorflow framework.
The U-Net baseline is trained for 65 epochs on a an NVIDIA Titan RTX GPU with 24GB RAM. We use Adam optimization~\cite{Adam} for model training.

\section{Improving the U-net baseline system}
\label{improve}
The improvement of the U-Net baseline focuses on three main aspects of a deep learning model: the loss function, the input quality and the network architecture.

\subsection{A combined loss function}
\label{loss}
We tackle the issue of class imbalance between event pixels and non-event pixels by using Focal loss~\cite{focal_loss}. Additionally, we apply IoU loss~\cite{iou_loss} to further improve the mIoU score within the segmentation task.
As a result, the final loss, referred to as the combined loss, is defined by combining Focal loss and IoU loss with equal weight.

\subsection{Input image quality enhancement}
\label{feature}
Feature engineering and augmentation are important tuning knobs for improving model performance. We therefore supplement the 14 original bands from the Landslide4Sense development set with 12 additional bands (bands 15 to 26), using methods methods as detailed in Table \ref{table:res_t0}.

\begin{itemize}
\item Bands 15 to 17 are generated by applying RGB normalization on bands B2, B3 and B4.
\item Bands 18 to 21 represent remote sensing indices (NDVI, NDMI, NBR) and a grayscale image.
\item Bands 22 and 23 are generated by applying Gaussian and median filters with kernel size of [10$\times$10].
\item Bands 24 and 25 are calculated from the image gradient (across length and width dimension).
\item Band 26 presents the result of using Canny edge detector.
\end{itemize}

\begin{table}[ht]
    \caption{Feature engineering: additional bands} 
    \vspace{-0.2cm}
    \centering
    \scalebox{0.85}{
    \begin{tabular}{ll} 
        \hline 
        \textbf{New band data}   &  \textbf{Formula / Method} \\
        \hline
        Band 15 to Band 17 &  $(x - x\_min)/(x\_max - x\_min)$ \\
        Band 18: NDVI & $(B8-B4)/(B8+B4)$\\ 
        Band 19: NDMI &  $(B8-B11)/(B8+B11)$\\ 
        Band 20: NBR &  $(B8-B12)/(B8+B12)$ \\ 
        Band 21:Gray &  $(B2+B3+B4)/3$ \\ 
        Band 22 to Band 23 & Gausian and Median filters\\
        Band 24 to Band 25 & Image gradients across length and width\\
        Band 26 & Canny Edge detector\\
        \hline 
    \end{tabular}
    }
    \vspace{-0.2cm}
    \label{table:res_t0} 
\end{table}

\subsection{U-Net backbone architecture improvement}
\label{architecture}
\begin{figure}[ht]
    \centering
    \includegraphics[width =0.6\linewidth]{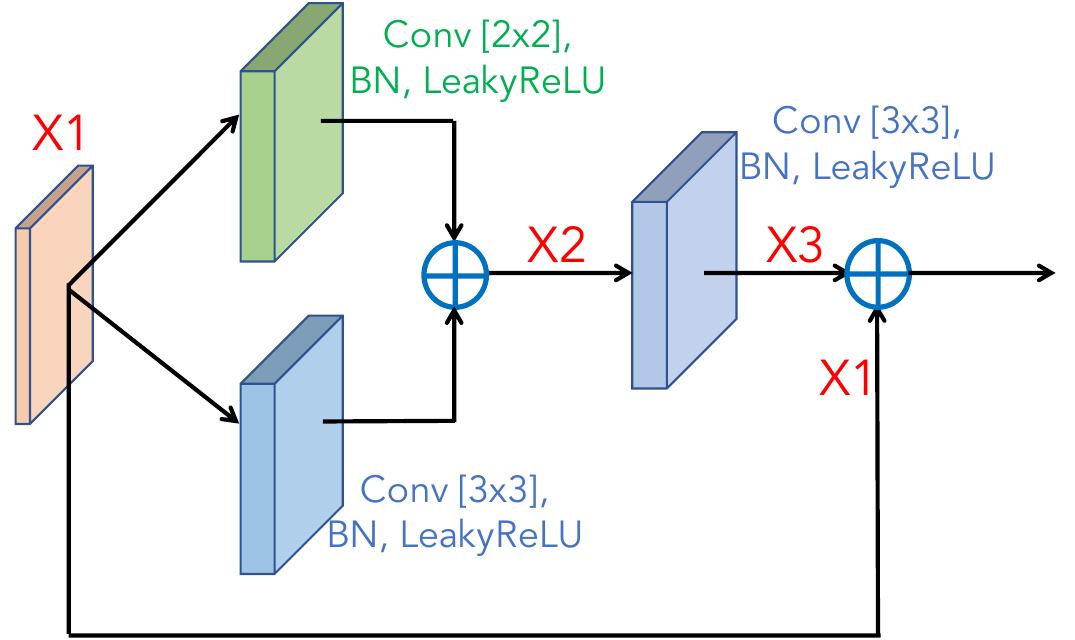}
   	\vspace{-0.2cm}
	\caption{Residual-convolutional layer.}
    \label{fig:f13}
    \vspace{-0.3cm}
\end{figure}

We propose three main improvements regarding the U-Net baseline architecture.
First, we suggest that multiple kernel sizes and a residual based architecture is more effective to capture distinct features of feature maps rather than a conventional convolutional layer.  
We therefore propose an architecture of a residual-convolutional layer (Res-Conv) which is used to replace the double convolution layer in both the downsample and upsample blocks.
Within the proposed residual-convolutional layer (Fig.~\ref{fig:f13}), the input feature map $\mathbf{X_1}$ is first learned by two convolutional layers with different kernels (e.g. Conv[2$\times$2] and Conv[3$\times$3]) before going through a BN layer, LeakyReLU layer and adding together to generate the feature map $\mathbf{X_2}$.
Then, the feature map $\mathbf{X_2}$ goes through a convolutional layer (Conv[3$\times$3]), BN layer, LeakyReLU layer to generate the feature map $\mathbf{X_3}$.
Finally, the feature map $\mathbf{X_3}$ is added with the input feature map $\mathbf{X_1}$ to create the final output of the proposed residual-convolutional sub-block.

\begin{figure}[ht]
    \vspace{-0.2cm}
    \centering
    \includegraphics[width =0.85\linewidth]{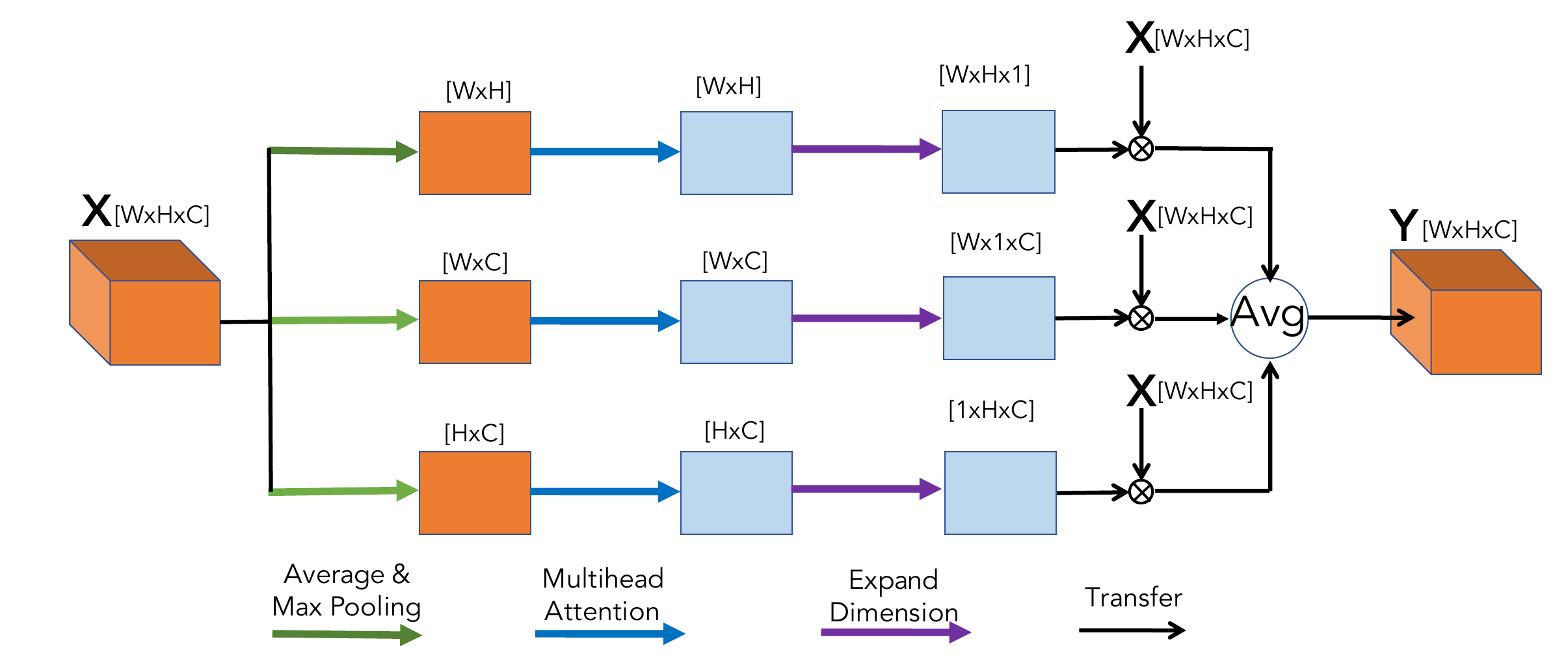}
   	\vspace{-0.2cm}
	\caption{Attention layer leveraging multi-head attention.}
    \label{fig:f14}
\end{figure}

The second improvement is to apply an attention layer after every convolutional layer in both the downsample and upsample blocks of the proposed U-Net baseline.
The attention weights generated by the proposed attention layer effectively support the neural network to focus on landslide regions on the feature maps in the network.  
We evaluate three types of attention schemes: SE~\cite{se_att} attention, CBAM~\cite{cbam_att} attention, and multi-head attention~\cite{multi-head_att}.
Both SE and CBAM are popular and widely used in literature.
Following the line of Le et al. (2023)~\cite{le2023robust}, we propose an additional multi-head attention based layer (Pro-Att) as follows: Given an input feature map $\mathbf{X}$ with a size of [W$\times$H$\times$C] where W, H, and C presents width, height, and channel dimensions, we reduce the size of feature map $\mathbf{X}$ across three dimensions using both max and average pooling layers (Fig.~\ref{fig:f14}). 
The multi-head attention scheme is then applied to each two-dimensional feature maps before multiplying with the original three-dimensional feature map $\mathbf{X}$.

The final improvement is inspired by applying an ensemble of multiple predicted masks with different resolutions to enhance the system performance.
In particular, instead of using only one head block to generate one predicted mask of 128$\times$128, we add two more head blocks to generate two other predicted masks: 256$\times$256 and 64$\times$64.
As a result, the final predicted result is obtained from an average of three predicted output masks. 
As we generate three predicted masks, three loss functions are applied for the learning process.

\section{Experimental results}
We first evaluate the effect of using the proposed combined loss function using the original images with with 14 bands only.
Both Focal loss and IoU loss achieve better performance than the Cross Entropy loss (Table~\ref{table:res_t1}).
The combination of Focal loss and IoU loss yields improvements of 1.22 in the F1-score and 1.13 in the mIoU score, respectively.

\begin{table}[ht]
    \caption{Effect of the combined loss function (using 80/20 splitting).} 
    \vspace{-0.2cm}
    \centering
    \scalebox{0.85}{
    \begin{tabular}{lcc} 
        \hline 
        \textbf{Networks \& Loss}   &  \textbf{F1 score}  &  \textbf{mIoU} \\       
        \hline
         U-Net  \& Cross Entropy (U-Net baseline) & 67.83 & 60.01\\
         U-Net  \& Focal Loss    & 68.28 & 60.37 \\
         U-Net  \& IoU Loss      & 68.20 & 60.23 \\
         \textbf{U-Net  \& Combined loss} & \textbf{69.05} & \textbf{61.14}\\
       \hline 
    \end{tabular}
    }
    \vspace{-0.3cm}
    \label{table:res_t1} 
\end{table}

As the proposed combined loss proved to be effective it was set as standard for further evaluation of the newly engineered features.
To assess the added value of the new features, the enhanced image input was trained with U-Net baseline using the combined loss. The use of the additional 12 bands leads to further performance improvements by 0.81 in F1-score and 0.62 in mIoU score compared with the U-Net baseline and combined loss (Table~\ref{table:res_t2}).

\begin{table}[ht]
    \caption{Effect of feature engineering (U-Net*: U-Net baseline with combined loss function and 80/20 splitting).} 
    \vspace{-0.2cm}
    \centering
    \scalebox{0.85}{
    \begin{tabular}{lcc} 
        \hline 
        \textbf{Band number}   &  \textbf{F1 score}  &  \textbf{mIoU} \\
        \hline
         U-Net* \& Original 14 bands                     &69.05 &61.14\\
         U-Net* \& Original 14 bands \& bands 15 to 17   &69.39 &61.22\\
         U-Net* \& Original 14 bands \& bands 15 to 21   &69.83 &60.97\\
         U-Net* \& Original 14 bands \& bands 15 to 23   &\textbf{69.96} &\textbf{61.76}\\
         U-Net* \& Original 14 bands \& bands 15 to 25   &69.91 &61.64\\
         U-Net* \& Original 14 bands \& bands 15 to 26   &68.54 &60.65\\
       \hline 
    \end{tabular}
    }
    \label{table:res_t2} 
\end{table}

We now evaluate the proposed multiple resolution heads, the proposed residual-convolutional layer, and the proposed attention layer.
To this end, we use the U-Net baseline, the full 23 band data and the combined loss.
All proposed techniques improve the U-Net model performance further (Table~\ref{table:res_t4}).
While the combination of multiple heads and the proposed attention layer achieves F1/mIoU scores of 71.45/63.05, the combination of multiple heads and the proposed residual-convolutional layer obtains the F1/mIoU scores of 72.07/63.45.

\begin{table}[ht]
    \caption{Effect of improving U-Net architecture (U-Net\ding{61}: U-Net baseline with combined loss function, 23 band data, 80/20 splitting)} 
    \vspace{-0.2cm}
    \centering
    \scalebox{0.85}{
    \begin{tabular}{lcc} 
        \hline 
        \textbf{Networks}   &  \textbf{F1 score}  &  \textbf{mIoU} \\       
        \hline
         U-Net\ding{61}                                 &69.96 &61.76\\ 
         U-Net\ding{61} \& Multiple heads               &70.45 &62.19\\ 
         U-Net\ding{61} \& Multiple heads \& CBAM Att   &70.82 &62.53\\
         U-Net\ding{61} \& Multiple heads \& SE Att     &71.26 &62.86\\
         U-Net\ding{61} \& Multiple heads \& Pro-Att    &71.45 &63.05\\
         U-Net\ding{61} \& Multiple heads \& Res-Conv  &\textbf{72.07} &\textbf{63.45}\\
       \hline 
    \end{tabular}
    }
    \vspace{-0.3cm}
    \label{table:res_t4} 
\end{table}

\begin{figure}[ht]
    \centering
    \includegraphics[width = 1.0\linewidth]{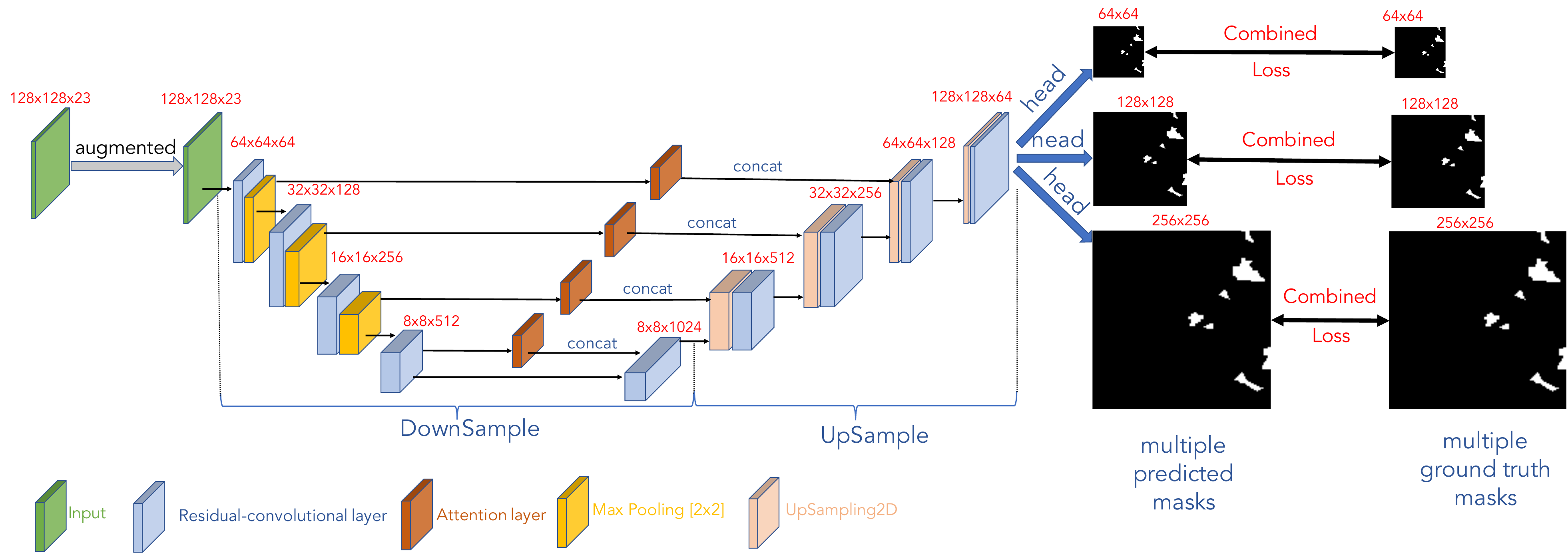}
   	\vspace{-0.2cm}
	\caption{Proposed optimal U-Net architecture for landslide detection and segmentation using remote sensing imagery.}
    \label{fig:f1}
\end{figure}

Given the effectiveness of using the combined loss function, the enhanced 23 band data, multi-resolution heads, the proposed Res-Conv layer and attention layers, we eventually configure the best U-Net network architecture (Fig.~\ref{fig:f1}).
We evaluate this network with 5-fold cross validation and compare it to the Landslide4Sense challenge baseline as well as the proposed U-Net baseline.
The best U-Net network achieves mIoU/F1 scores of 76.07/84.07, thereby outperforming the Landslide4Sense challenge baseline and the proposed U-Net baseline (Table~\ref{table:res_t4}). The best U-Net network architecture also performs the lower trainable parameters compared to the other networks.

\begin{table}[ht]
    \caption{Performance comparison between the Landslide4Sense baseline, the proposed U-Net baseline, and the best U-Net network architecture on the Landslide4Sense development set with 5-fold cross validation} 
    \vspace{-0.2cm}
    \centering
    \scalebox{0.85}{
    \begin{tabular}{lccc} 
        \hline 
        \textbf{Networks}   &  \textbf{F1 scores}  &  \textbf{mIoU} &\textbf{Parameters (M)}\\       
        \hline
         Landslide4Sense baseline~\cite{chal_baseline}       &77.19 &68.64 &29.8\\ 
         Proposed U-Net baseline                 &73.51 &67.20 &31.0\\ 
         The best U-Net based network   &\textbf{84.07} & \textbf{76.07} &\textbf{24.8} \\
       \hline 
    \end{tabular}
    }
    \vspace{-0.3cm}
    \label{table:res_t4} 
\end{table}

\section{Conclusion}

We have presented a U-Net based deep neural network for landslide detection and segmentation from the remote sensing imagery.
We consider and evaluate the effects of improvements of feature engineering, network architecture, and loss functions, and illustrate corresponding improvements in overall network performance.
By conducting extensive experiments, we successfully developed an U-Net neural network which achieves an F1-score of 84.07 and an mIoU score of 76.07 on the benchmark Landslide4Sense development set.
Our proposed system clearly outperforms the Landslide4Sense baseline by improving the F1-score by 6.88 and the and mIoU score by 7.43, respectively.

\section*{ACKNOWLEDGMENTS}
The research leading to this publication was partially carried out within the gAia project. The gAia project is funded by the KIRAS program of the Austrian Research Promotion Agency (FFG) and the  Federal Ministry of Agriculture, Regions and Tourism (BMLRT) under grant no. FO999886369.

\newpage

\addtolength{\textheight}{-11cm}   

\bibliographystyle{IEEEbib}
\bibliography{refs}
\end{document}